\documentclass{article}

\usepackage{PRIMEarxiv}

\usepackage[utf8]{inputenc} 
\usepackage[T1]{fontenc}    
\usepackage{hyperref}       
\usepackage{url}            
\usepackage{booktabs}       
\usepackage{amsfonts}       
\usepackage{nicefrac}       
\usepackage{microtype}      
\usepackage{lipsum}
\usepackage{fancyhdr}       
\usepackage{graphicx}       
\graphicspath{{media/}}     

\usepackage{tabularx}
\usepackage[outdir=./]{epstopdf}
\usepackage{graphics}
\usepackage{verbatim}

\usepackage{multirow}
\usepackage{multicol}
\usepackage[table,xcdraw]{xcolor}
\usepackage{boldline}

\usepackage{cuted}
\usepackage{capt-of}

\usepackage{amsmath}
\DeclareMathOperator*{\argmin}{arg\,min}

\title{Joint Video Enhancement with Deblurring, Super-Resolution, and Frame Interpolation Network}

\author{
  Giyong Choi, HyunWook Park \\
  School of Electrical Engineering \\
  KAIST \\
  Daejeon \\
  \texttt{\{gychoi92, hwpark\}@kaist.ac.kr} \\
}

\begin{document}
\maketitle

\begin{abstract}
Video quality is often severely degraded by multiple factors rather than a single factor. These low-quality videos can be restored to high-quality videos by sequentially performing appropriate video enhancement techniques. However, the sequential approach was inefficient and sub-optimal because most video enhancement approaches were designed without taking into account that multiple factors together degrade video quality. In this paper, we propose a new joint video enhancement method that mitigates multiple degradation factors simultaneously by resolving an integrated enhancement problem. Our proposed network, named DSFN, directly produces a high-resolution, high-frame-rate, and clear video from a low-resolution, low-frame-rate, and blurry video. In the DSFN, low-resolution and blurry input frames are enhanced by a joint deblurring and super-resolution (JDSR) module. Meanwhile, intermediate frames between input adjacent frames are interpolated by a triple-frame-based frame interpolation (TFBFI) module. The proper combination of the proposed modules of DSFN can achieve superior performance on the joint video enhancement task. Experimental results show that the proposed method outperforms other sequential state-of-the-art techniques on public datasets with a smaller network size and faster processing time.
\end{abstract}


\section{Introduction}

The demands for the high-quality video have been steadily growing over the past decades. The high-quality video usually requires the larger number of pixels for high spatial resolution, more frames per second for high temporal resolution, and a faster shutter speed for clear frames. However, the videos obtained from the cameras that cannot fully support these technical specifications may have low qualities. To meet the demand for high-quality video, these low-quality videos can be improved by applying appropriate video enhancement techniques.

Many video enhancement techniques have been introduced to enhance the video quality. The most widely used video enhancement technologies are video super-resolution (VSR) (i.e. improving spatial resolution), video frame interpolation (VFI) (i.e. increasing temporal resolution), and video deblurring (VD) (i.e. removing motion blurs). In VSR, high-resolution patches can be produced by using one or more low-resolution patches \cite{6haris2019recurrent,29liao2015video,30jo2018deep,31kappeler2016video,32caballero2017real}. VFI increases the frame rate by creating and inserting the intermediate frames into the video sequence \cite{12niklaus2017video,13niklaus2017video,14niklaus2018context,15bao2019depth}. In addition, VD makes frames clearer by removing motion blurs on input blurry frames \cite{40zhou2019spatio,41wang2019edvr,42pan2020cascaded}. 

Each of these video enhancement techniques has achieved good performance in reducing individual degradation factors rather than multiple degradation factors. In practice, however, videos are often degraded simultaneously by several factors. When videos are degraded by multiple factors, the videos need to be processed taking into account multiple degradation factors together. The simplest solution is to apply multiple video enhancement techniques sequentially. For instance, to restore a high-resolution and high-frame-rate clear video from a low-resolution and low-frame-rate blurry video, proper enhancement techniques, i.e. VD, VSR, and VFI in this case, can be sequentially performed. However, this sequential approach is sub-optimal and has some vital problems. Most of all, existing video enhancement techniques may not be suitable because they were originally designed only for a single degradation factor. When an enhancement technique deals with the particular degradation factor, other degradation factors would interfere with the enhancement, which results in low performance. Besides, when several enhancement techniques are applied sequentially, the errors that occur in the previous step can be propagated to the next step, so it causes error accumulation and poor performance as well. Lastly, it would have relatively high computational complexity and be inefficient.

To overcome these problems, more recently, several methods that jointly deal with multiple degradation factors have been proposed. Shen \textit{et al}. \cite{43shen2020blurry} formulated a unified degradation model that performed both video deblurring and frame interpolation simultaneously to successfully enhance low-frame-rate blurry video. In addition, in \cite{44haris2020starnet} and \cite{45xiang2020zooming}, both spatial and temporal resolution of the video are improved at the same time, which restored a high-resolution and high-frame-rate video from a low-resolution and low-frame-rate video. These joint video enhancement methods allow optimal and efficient restoration of video degraded by two factors.

However, videos are often severely degraded by more than two factors. The videos, especially captured by hand-held cameras, may have a low spatial resolution and low frame rate, and have motion blurs. Even though several joint approaches were recently introduced in \cite{43shen2020blurry,44haris2020starnet,45xiang2020zooming}, they were not appropriate to fully handle these low-resolution, low-frame-rate, and blurry videos. To restore these videos, we newly propose a joint video enhancement technique that directly restores a high-resolution, high-frame-rate, and clear video from a low-resolution, low-frame-rate, and blurry video. In other word, the proposed method deals with three enhancement techniques of VSR, VFI, and VD at the same time. Because the proposed method resolves an integrated enhancement problem, the severely degraded videos can be restored effectively.

\begin{figure*}
\begin{center}
   \includegraphics[width=\linewidth]{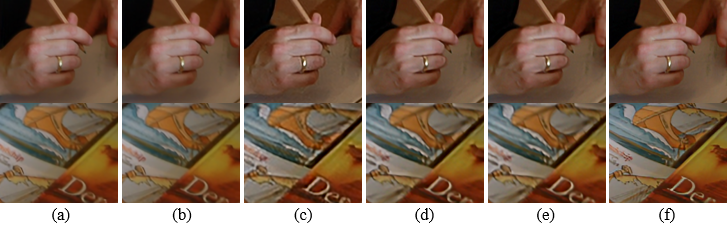}
\end{center}
\vspace{-0.5cm}
   \caption{Joint video enhanced frames from various methods on a real blurry dataset \cite{26cho2012video}. The various methods are (a) CDVD+STAR, (b) CDVD+Zooming-SloMo, (c) BIN+RBPN, (d) STFAN+STAR, (e) STFAN+Zooming-SloMo, and (f) the proposed method.}
\label{fig: real-world}
\end{figure*}

Our proposed Deblurring, Super-resolution, and Frame interpolation Network (DSFN) performs deblurring and up-scaling of input frames and creates intermediate frames between input adjacent frames. To effectively perform the multiple enhancement tasks, DSFN is designed elaborately with four modules: encoder, joint deblurring and super-resolution (JDSR) module, triple-frame-based frame interpolation (TFBFI) module, and decoder. The proposed JDSR module directly produces a deblurred and super-resolved feature map at once from a feature map obtained by the encoder. In the meantime, the proposed TFBFI module also takes the feature map from the encoder, and generates feature maps which correspond to the intermediate frames. The enhanced feature maps are then reconstructed into the desired high-resolution images. The proper combination of the proposed modules in DSFN can achieve superior performance on the joint video enhancement task, which is shown in Figure~\ref{fig: real-world}.

\begin{figure}
\begin{center}
   \includegraphics[width=0.6\linewidth]{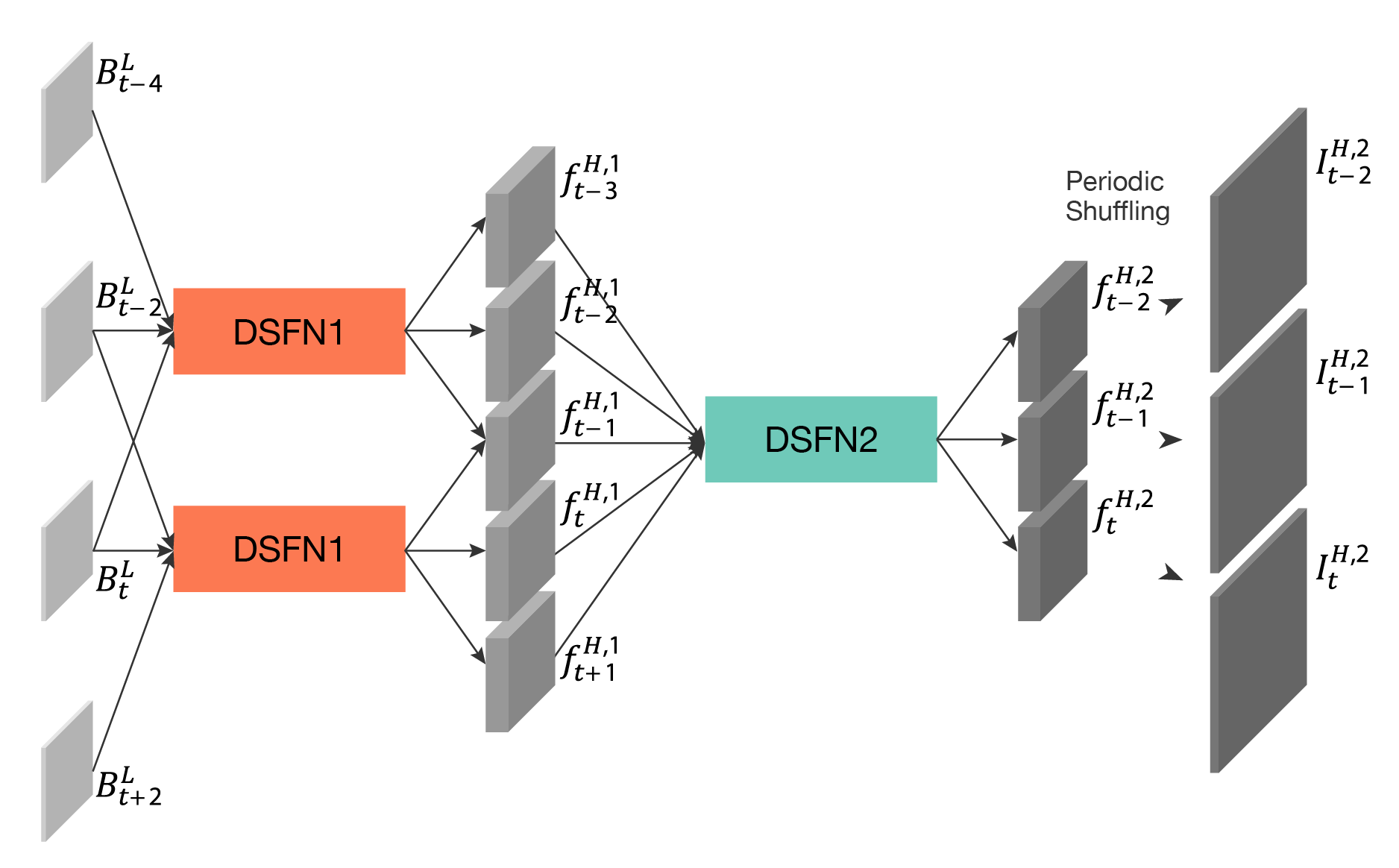}
\end{center}
\vspace{-0.5cm}
   \caption{Architecture of the proposed two-level DSFNs. The DSFN of the same color shares the network parameters.}
\label{fig:two-level}
\end{figure}

To restore severely distorted images and to further boost the performance, we employ two-level DSFNs as shown in Figure~\ref{fig:two-level}. The second level DSFN additionally refines the enhanced outputs of the previous DSFN. Rather than simply increasing the size of the network, using the multi-level architecture allows more effective learning by minimizing a loss of intermediate outputs between the stages. Experiments have shown that the proposed method has the superior performance compared to other sequential methods that are combinations of state-of-the-art (SOTA) video enhancement techniques on both public datasets: Adobe240 \cite{36su2017deep} and GOPRO \cite{GOPRO_Nah_2017_CVPR}. In addition, our approach performs the multiple tasks simultaneously on a relatively small network and short processing time.

The main contributions of the proposed method are described as follows: (1) The novel joint video enhancement network is introduced, which performs the three tasks of VD, VSR, and VFI, simultaneously. (2) The proposed network, named DSFN, effectively performs deblurring and up-scaling of input frames by the proposed JDSR module and creates intermediate frames between input adjacent frames by the proposed TFBFI module. The integrated enhancement model is formulated to achieve optimal joint video enhancement performance. (3) In our experiments, the proposed method achieved the superior performance compared to other sequential approaches that consist of multiple SOTA methods. While achieving high performance, the network size is small and the processing time is fast.

The rest of this paper is organized as follows. In Section \ref{sec: Related Works}, previous video enhancement techniques are introduced. Our approach is presented in Section \ref{sec: The Proposed Method}. Implementation and network training are explained in Section \ref{sec: Implementations and Training}. In Section \ref{sec: Experimental Results}, experiment results and discussions are presented. Finally, the paper is concluded in Section \ref{sec: Conclusion}.

\section{Related Works}
\label{sec: Related Works}

\subsection{Video Super-Resolution}

Unlike single-image super-resolution \cite{1dong2015image,2kim2016accurate,3lim2017enhanced,4ledig2017photo,5shi2016real}, which produces a high-resolution image from a single low-resolution input image, video super-resolution (VSR) has the advantage of utilizing information from neighbor frames to increase the resolution of a target frame. To utilize the information of neighbor frames, several VSR methods \cite{29liao2015video,30jo2018deep,31kappeler2016video,32caballero2017real} employed the concatenated low-resolution patches as the inputs. In these methods, the networks extracted appropriate features from the concatenated inputs to improve the resolution of the target frame. On the other hand, some VSR methods adopted a recurrent scheme. In \cite{33huang2015bidirectional,34tao2017detail,35sajjadi2018frame,isobe2020video}, the networks utilized the previous restored frame and the current input frame to improve the resolution of the current frame, where the previous frame was restored in the previous step. In addition, Haris \textit{et al}. \cite{6haris2019recurrent} proposed a new VSR method that employed a recurrent encoder-decoder module to merge spatial and temporal contexts from multiple frames. In \cite{fuoli2019efficient}, a recurrent latent space propagation (RLSP) algorithm, which introduced high-dimensional latent states, was proposed to implicitly propagate temporal information. Chan \textit{et al}. proposed the general and concise pipeline for VSR (BasicVSR \cite{chan2021basicvsr} and its enhanced version BasicVSR++ \cite{chan2022basicvsr++}) with attractive improvements in speed and performance.

\subsection{Video Frame Interpolation}

In video frame interpolation (VFI), object motions between successive frames should be considered elaborately to generate intermediate frames that maintain temporal continuity with the input frames. Conventional frame rate up-conversion (FRUC) methods \cite{7choi2000new,8kang2010dual,9wang2010frame,10dikbas2012novel,11choi2018triple} divided an input frame into blocks and employed block matching algorithms (BMAs) that estimated motion vectors for each block to compensate object motions for interpolated frames. Recently, plenty of deep-learning-based VFI methods have been proposed. In \cite{12niklaus2017video} and \cite{13niklaus2017video}, 2-D kernel was adaptively estimated for each pixel by using convolutional neural networks (CNNs). Every pixel of an intermediate frame was obtained by convolving the input patches with the estimated kernels. On the other hand, some VFI methods \cite{14niklaus2018context,15bao2019depth,16liu2019deep,17jiang2018super,20niklaus2020softmax} estimated optical flow maps between consecutive input frames and then created the intermediate frames by warping input frames in accordance with the optical flow maps. To produce more sophisticated intermediate frames, the additional information such as contextual information \cite{14niklaus2018context} and depth information \cite{15bao2019depth} was also utilized. In \cite{18meyer2018phasenet}, the interpolated frame was constructed from its phase-based representations that were estimated by PhaseNet, which was more resistant to changes in brightness. To handle the complex motions in videos, Lee \textit{et al}. \cite{19lee2020adacof} proposed a new warping module named Adaptive Collaboration of Flows (AdaCoF), which had a high degree of freedom.

\subsection{Video Deblurring}

Conventional video deblurring methods \cite{23hyun2015generalized,24ren2017video,25sroubek2011robust} formulated the blur models, which estimated a blur kernel to represent the relation between blurry frames and latent frames, and removed motion blurs on the frame by solving an inverse problem. In the aggregation-based video deblurring approaches \cite{26cho2012video,27delbracio2015hand,28delbracio2015burst}, blurry frames could be enhanced by borrowing sharp pixels from its neighbor frames because blurry pixels in a certain frame may appear sharp in other frames. Recently, many video deblurring methods that employed deep neural networks have been proposed \cite{36su2017deep,37hyun2017online,38wieschollek2017learning,39nah2019recurrent,40zhou2019spatio}. Su \textit{et al}. \cite{36su2017deep} proposed a new video deblurring method that employed CNN with stacked blurry frames. In \cite{40zhou2019spatio}, input blurry frames were deblurred by adaptive alignment and deblurring filters using a recurrent scheme. Besides, Pan \textit{et al}. \cite{42pan2020cascaded} proposed a simple and effective video deblurring method that generated latent frames from blurry inputs by using frame alignment and a temporal sharpness prior.

\subsection{Joint Video Enhancement}

More recently, joint methods that performed multiple video enhancement tasks concurrently have been proposed. Shen \textit{et al}. \cite{43shen2020blurry} proposed a blurry video frame interpolation (BIN) method that jointly performed the video deblurring and frame interpolation. In \cite{43shen2020blurry}, the backbone network took two blurry frames as inputs and produced a clear intermediate frame. To effectively exploit the temporal information, they adopted the recurrent module, i.e. ConvLSTM units \cite{xingjian2015convlstm}, to propagate the frame information over time. In addition, space-time super-resolution (ST-SR) approaches, which generated a high-resolution and high-frame-rate video from a low-resolution and low-frame-rate video, were proposed. Haris \textit{et al}. \cite{44haris2020starnet} proposed a space-time-aware multi-resolution network (STARnet) to create not only up-scaled high-resolution frames but also a high-resolution intermediate frame. In \cite{45xiang2020zooming}, high-resolution slow-motion videos could be obtained from low-resolution videos by employing a novel deformable ConvLSTM.

\begin{figure*}
\begin{center}
   \includegraphics[width=\linewidth]{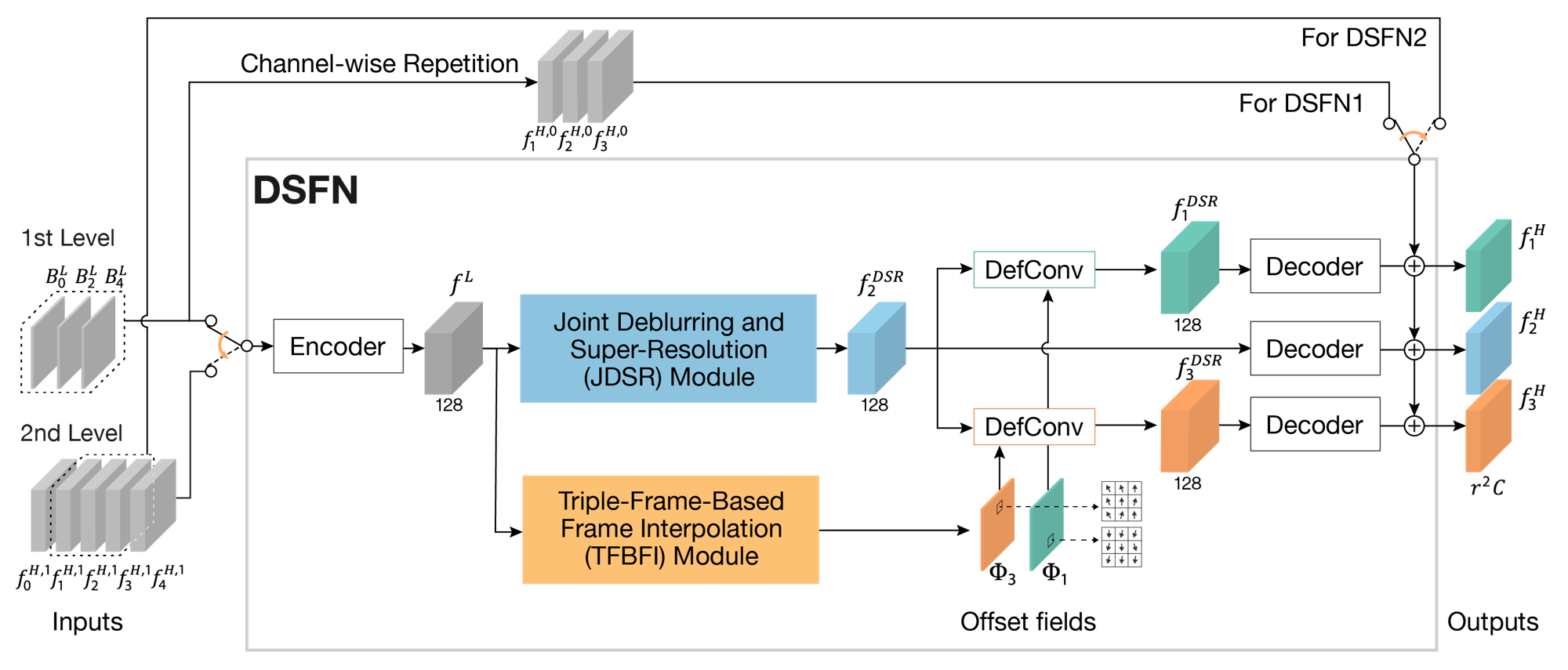}
\end{center}
\vspace{-0.5cm}
   \caption{Architecture of the proposed DSFN.}
\label{fig:DSFN}
\end{figure*}

\section{The Proposed Method}
\label{sec: The Proposed Method}

\subsection{Overview}

Given a low-resolution and low-frame-rate blurry video sequence, $\mathbf{B}^{L}=\left \{ B_{2t}^{L} \right \}_{t=0}^{n}$, we aim to generate the corresponding high-resolution and high-frame-rate clear video sequence as close as the ground-truth, $\mathbf{I}^{GT}=\left \{ I_{t}^{GT} \right \}_{t=1}^{2n-1}$. The proposed Deblurring, Super-resolution, and Frame interpolation Network (DSFN) performs deblurring and up-scaling of input frames and creates intermediate frames between the input adjacent frames by using the integrated enhancement model as follows:
\begin{equation} \label{}
\theta^{*}=\argmin_{\theta} \rho(\mathcal{N}_{DSFN}(\mathbf{B}^{L};\theta),\mathbf{I}^{GT})
\end{equation}
where $\rho$ is a loss function and $\mathcal{N}_{DSFN}$ denotes the proposed network with a set of optimized parameters of $\theta^{*}$. The proposed DSFN consists of four modules: encoder, joint deblurring and super-resolution (JDSR) module, triple-frame-based frame interpolation (TFBFI) module, and decoder as shown in Figure~\ref{fig:DSFN}. The proper combination of these modules can effectively solve the joint video enhancement problem. 

In addition, we adopt two-level DSFNs of $\textup{DSFN}1$ and $\textup{DSFN}2$ to handle severely distorted images and to further improve the performance. As illustrated in Figure~\ref{fig:two-level}, given consecutive blurry frames $(B_{t-4}^{L},B_{t-2}^{L},B_{t}^{L},B_{t+2}^{L})$, the deblurred and super-resolved feature maps $(f_{t-3}^{H,1},f_{t-2}^{H,1},f_{t-1}^{H,1},f_{t}^{H,1},f_{t+1}^{H,1})$ are produced by the two $\textup{DSFN}1$s. 
Like $\textup{DSFN}1$, $\textup{DSFN}2$ also needs to receive a total of three feature maps $(f_{t-3}^{H,1},f_{t-1}^{H,1},f_{t+1}^{H,1})$ based on the original frame rate. It is worth noting that the enhanced feature maps, $(f_{t-2}^{H,1}$ and $f_{t}^{H,1})$, are also entered as inputs to the second level for additional performance gains. 
Then, these output feature maps are further enhanced to $(f_{t-2}^{H,2},f_{t-1}^{H,2},f_{t}^{H,2})$ by $\textup{DSFN}2$. The output feature maps can be converted to the desired high-resolution images by a periodic shuffling operation. 
Consequently, the enhanced frames $(I_{t-2}^{H,2},I_{t-1}^{H,2},I_t^{H,2})$ are produced by the two-level DSFNs given input blurry frames $(B_{t-4}^{L},B_{t-2}^{L},B_{t}^{L},B_{t+2}^{L})$.
Detailed explanations of each module of the DSFN and the proposed method are provided in the next subsections.

\subsection{DSFN: Encoder}

Firstly, the encoder of DSFN extracts appropriate features $f^{L}$ from consecutive blurry inputs as follows:
\begin{equation} \label{}
f^{L}=\mathcal{N}_{enc}(B_{0}^{L},B_{2}^{L},B_{4}^{L};\theta_{enc,1})~~\textup{for}~\textup{DSFN}1,
\end{equation}
\begin{equation} \label{}
f^{L}=\mathcal{N}_{enc}(f_{0}^{H,1},f_{1}^{H,1},f_{2}^{H,1},f_{3}^{H,1},f_{4}^{H,1};\theta_{enc,2})~~\textup{for}~\textup{DSFN}2,
\end{equation}
where $\mathcal{N}_{enc}$ denotes the encoder with a set of parameters of $\theta_{enc,1}$ and $\theta_{enc,2}$ for $\textup{DSFN}1$ and $\textup{DSFN}2$, respectively. The inputs of the encoder are blurry inputs for $\textup{DSFN}1$ and the output feature maps of $\textup{DSFN}1$ for $\textup{DSFN}2$. The extracted feature map $f^{L}$ contains the appropriate information of blurry inputs and it is utilized in both JDSR and TFBFI modules.

\subsection{DSFN: Joint Deblurring and Super-Resolution (JDSR) Module}

Given the feature map $f^{L}$ from the encoder, the proposed JDSR module performs deblurring and up-scaling at once for the central frame of the inputs. Because not only the central frame but also its neighbor frames are encoded in the feature map $f^{L}$, we can restore the central frame more effectively with a lot of relevant information. Through the JDSR module, a deblurred and super-resolved feature map, $f_{2}^{DSR}$, which corresponds to the central frame is obtained from the feature map, $f^{L}$, as follows:
\begin{equation} \label{}
f_{2}^{DSR}=\mathcal{N}_{DSR}(f^{L};\theta_{DSR}),
\end{equation}
where $\mathcal{N}_{DSR}$ denotes the JDSR module with a set of parameters $\theta_{DSR}$. In contrast to $f^{L}$, the enhanced feature map $f_{2}^{DSR}$ contains information of the central frame only.

Many VSR methods usually employ a simple filter, such as bicubic interpolation, to up-scale input frames before adopting a super-resolution operation. However, performing the filtering operation with a handcrafted filter is sub-optimal and increases computational complexity.
To achieve better performance and efficiency, the proposed JDSR module treats the inputs in the original input scale. The feature map $f_{2}^{DSR}$ has the same size as $f^{L}$, and contains the deblurred and high-resolution information in channel dimension, rather than spatial dimension. This allows the JDSR module to effectively perform not only super-resolution but also deblurring at the same time. The desired high-resolution images are obtained by simply applying periodic shuffling to the feature maps. Besides, the following TFBFI module can efficiently provide offset fields for frame interpolation in the input scale.

\subsection{DSFN: Triple-Frame-Based Frame Interpolation (TFBFI) Module}

To up-convert the frame rate of the videos, the proposed TFBFI module generates two intermediate frames between three input keyframes. To produce the intermediate frames that maintain temporal continuity with the input frames, the movement of objects between successive frames should be considered elaborately. In the TFBFI module, therefore, two offset fields which infer motion of objects are predicted from $f^{L}$ as follows:
\begin{equation} \label{two_off}
(\Phi_{1},\Phi_{3})=\mathcal{N}_{FI}(f^{L};\theta_{FI}),
\end{equation}
where $\mathcal{N}_{FI}$ denotes the TFBFI module with a set of parameters $\theta_{FI}$. $\Phi_{1}$ and $\Phi_{3}$ are the offset fields that are used to generate the warped feature maps of $f_{1}^{DSR}$ and $f_{3}^{DSR}$, respectively.

Then, the feature map of the intermediate frame can be obtained by warping the deblurred and super-resolved feature map $f_{2}^{DSR}$ with deformable convolution \cite{22dai2017deformable} in accordance with the obtained offset fields as follows:
\begin{equation} \label{}
f_{i}^{DSR}=DefConv(f_{2}^{DSR},\Phi_{i})~~\textup{for}~i=1~\textup{or}~3,
\end{equation}
where $DefConv(\cdot,\cdot)$ denotes the deformable convolution with the offset field $\Phi_{i}$.

In \cite{41wang2019edvr} and \cite{tian2020tdan}, the neighbor frames are warped to the target frame by using deformable convolution \cite{22dai2017deformable} with the obtained offset values. However, they only produce one offset field at a time, so they need to estimate multiple offset fields, where the number of offset fields is equal to the number of warped frames. On the other hand, the proposed TFBFI module estimates two offset fields concurrently as in Eq.~(\ref{two_off}). Therefore, we can obtain multiple intermediate frames more efficiently in the TFBFI module.

In addition, other video frame interpolation methods, such as \cite{12niklaus2017video,13niklaus2017video,18meyer2018phasenet}, and \cite{19lee2020adacof}, utilized only two keyframes to generate intermediate frames. On the contrary, the proposed TFBFI module can achieve better frame interpolation performance because three consecutive keyframes are employed to provide more information.

\subsection{DSFN: Decoder}

Through the proposed JDSR and TFBFI modules, we can obtain the deblurred and super-resolved feature map, $f_{2}^{DSR}$, and two intermediate feature maps, $f_{1}^{DSR}$ and $f_{3}^{DSR}$. Their spatial dimension is the same as the input frames of $H\times W$, but the high-resolution information is contained in multiple channels. In the decoder, the number of channels of the obtained feature maps are adjusted to $r^{2}C$, where $r$ is a up-scaling factor and $C$ is the number of input image channels. Given the feature maps $f_{i}^{DSR}$ for $i\in\{1,2,3\}$, the decoder produces the high-resolution feature map $f_{i}^{H}$ as follows:
\begin{equation} \label{}
f_{i}^{H,1}=\mathcal{N}_{dec}(f_{i}^{DSR};\theta_{dec,1})+f_{i}^{H,0}~~\textup{for}~\textup{DSFN}1,
\end{equation}
\begin{equation} \label{}
f_{i}^{H,2}=\mathcal{N}_{dec}(f_{i}^{DSR};\theta_{dec,2})+f_{i}^{H,1}~~\textup{for}~\textup{DSFN}2,
\end{equation}
where $\mathcal{N}_{dec}$ denotes the decoder with a set of parameters of $\theta_{dec,1}$ and $\theta_{dec,2}$ for $\textup{DSFN}1$ and $\textup{DSFN}2$, respectively. 
In $\textup{DSFN}1$, the input adjacent keyframes are averaged and then repeated channel by channel to generate $f_{1}^{H,0}$ and $f_{3}^{H,0}$ as follows:
\begin{equation} \label{}
\tilde{B}^{L}_{1}=\frac{1}{2}(B^{L}_{0}+B^{L}_{2}), ~~\tilde{B}^{L}_{3}=\frac{1}{2}(B^{L}_{2}+B^{L}_{4}),
\end{equation}
\begin{equation} \label{}
{f}^{H,0}_{1}=Rep(\tilde{B}^{L}_{1}),
~~{f}^{H,0}_{2}=Rep({B}^{L}_{2}),
~~{f}^{H,0}_{3}=Rep(\tilde{B}^{L}_{3}),
\end{equation}
where $\tilde{B}^{L}_{1}$ and $\tilde{B}^{L}_{3}$ are averaged frames and $Rep(\cdot)$ denotes the channel-wise repetition. All decoders share their network parameters for each level.

\subsection{Reconstruction to High-Resolution Images}

As in \cite{5shi2016real}, the feature map of $H\times W\times r^{2}C$ can be rearranged to the high-resolution image of $rH\times rW\times C$ by a periodic shuffling operator. For $i\in\{1,2,3\}$, the desired high-resolution images are obtained from the high-resolution feature map $f_{i}^{H}$ as follows:
\begin{equation} \label{}
I_{i}^{H}=PS(f_{i}^{H}),
\end{equation}
where $PS$ is a periodic shuffling introduced in \cite{5shi2016real}. In the proposed DSFN, the desired outputs can be efficiently obtained because video enhancement tasks are performed on the feature maps with a low dimension, rather than the images with a high dimension.

\subsection{Multi-Level DSFNs}

A single DSFN could achieve acceptable joint video enhancement performance. However, when the inputs are quite blurry or severely distorted, the single DSFN may be insufficient to deal with their degradations. In \cite{17jiang2018super} and \cite{41wang2019edvr}, they adopted a multi-level strategy to achieve better results. Likewise, to handle the inputs that are blurry or severely distorted and to further boost the performance, the proposed method adopts two-level DSFNs as shown in Figure~\ref{fig:two-level}. Because a loss can be given to intermediate outputs between the stages, the multi-level architecture can achieve more effective learning.

In the proposed method, the second level DSFN additionally refines the enhanced outputs of the first level DSFN. In the first level, four consecutive low-resolution and blurry frames $(B_{t-4}^{L},B_{t-2}^{L},B_{t}^{L},B_{t+2}^{L})$ are employed as the inputs. Three out of four input frames are then fed into each $\textup{DSFN}1$. Two $\textup{DSFN}1$s of the first level share their parameter values. Two $\textup{DSFN}1$s of the first level generate two deblurred and super-resolved feature maps $(f_{t-2}^{H,1},f_{t}^{H,1})$ and three intermediate feature maps $(f_{t-3}^{H,1},f_{t-1}^{H,1},f_{t+1}^{H,1})$. The duplicated frames frames (i.e., $f_{t-1}^{H,1}$ in Figure~\ref{fig:two-level}) from two $\textup{DSFN}1$s are averaged. These enhanced outputs from the first level are further enhanced in the second level $\textup{DSFN}2$. Unlike the first level, the intermediate outputs are used as the inputs to the $\textup{DSFN}2$ to utilize more information. Thus, $\textup{DSFN}2$ receives five feature maps and produces further enhanced output feature maps $(f_{t-2}^{H,2},f_{t-1}^{H,2},f_{t}^{H,2})$. The final desired high-resolution images $(I_{t-2}^{H,2},I_{t-1}^{H,2},I_{t}^{H,2})$ are then obtained by periodic shuffling of the output feature maps from $\textup{DSFN}2$.

\section{Implementations and Training}
\label{sec: Implementations and Training}

\subsection{Dataset}

To train the proposed DSFN, we employed the Adobe240 dataset \cite{36su2017deep} which contained real-world videos at 240 fps with the resolution of $640\times352$. Because the video frames in \cite{36su2017deep} were obtained at a very high frame rate and with a fast shutter speed, they could be regarded as the ground-truth clear frames. Then, we generated their corresponding blurry frames by accumulating several consecutive frames. As \cite{43shen2020blurry} did, we accumulated 11 consecutive frames to generate a blurry frame, and down-sampled in time, which resulted in blurry videos at 30 fps. In addition, we performed several data augmentation techniques for each training sequence, such as random cropping to $256\times256$ patches, random horizontal or vertical flipping, color jittering, and adding Gaussian random noise of $N(0,0.01)$ to the images whose range was [0,1].

\subsection{Network Training}

With the training dataset, the proposed network was trained to produce the deblurred and up-scaled central frame and the intermediate frames from blurry inputs. In our experiments, we adopted a $4\times$ up-scaling factor (i.e. $r=4$) for fair comparison study. Therefore, input patches were randomly cropped from the training sequence with the size of $256\times256$, and they were reduced to $1/4$ size with the bicubic kernel. The size of input patches was then recovered to the original spatial resolution by the proposed network. To effectively train the proposed network which consisted of the first level and second level DSFNs, we applied the mean absolute error (MAE) loss function to the first-level outputs as well as the final outputs as follows:
\begin{equation} \label{}
\mathcal{L}=\frac{1}{5}\sum_{i\in\Omega_{1}}\left \| I_{i}^{H,1}-I_{i}^{GT} \right \|_{1}+\frac{1}{3}\sum_{i\in\Omega_{2}}\left \| I_{i}^{H,2}-I_{i}^{GT} \right \|_{1},
\end{equation}
where $\Omega_{1}$ and $\Omega_{2}$ are indices of output frames of $\textup{DSFN}1$ and $\textup{DSFN}2$, respectively, and $I_{i}^{H,1}$ is the periodically shuffled images from $f_{i}^{H,1}$. Besides, $I_{i}^{GT}$ is the corresponding ground-truth frame and $|| \cdot ||_{1}$ denotes the L1 norm. The entire network is jointly trained in an end-to-end manner with the above loss function.

\subsection{Implementation details}

\begin{figure}
\begin{center}
    \includegraphics[width=0.6\linewidth]{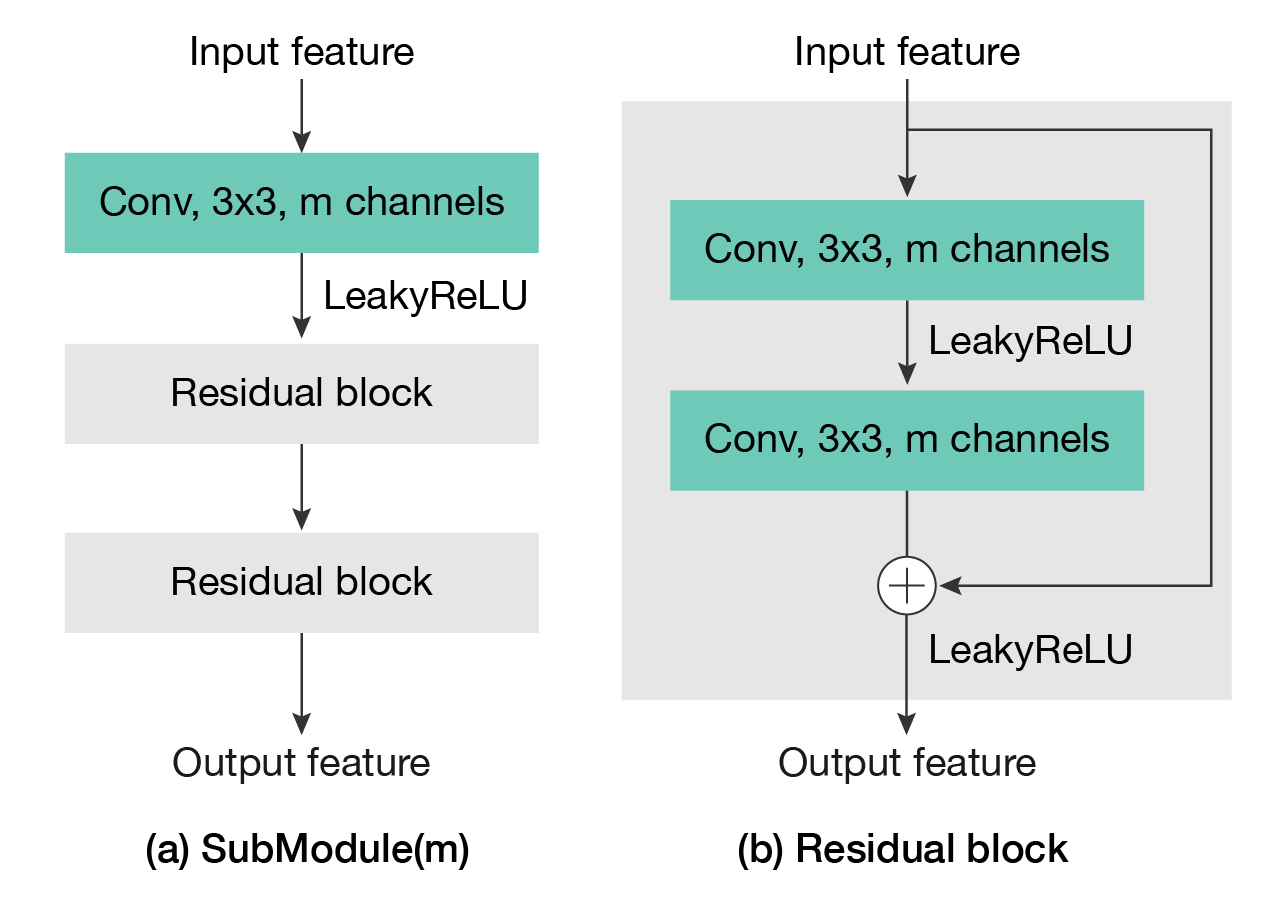}
\end{center}
\vspace{-0.5cm}
   \caption{Architecture of (a) SubModule with m channels and (b) residual block.}
\label{fig:submodule}
\end{figure}

Every module in the proposed DSFN consists of multiple convolutional layers and residual blocks \cite{resblock2016deep}, which are a combination of two convolutional layers and identity mapping. In addition, LeakyReLU with a slope of $\lambda=0.1$ is employed as the activation function. Let $\textup{SubModule}(m)$ be composed of one convolutional layer and two residual blocks with $m$ output channels, which is shown in Figure~\ref{fig:submodule}. Then, each module of DSFN can be represented as follows: $\textup{SubModule}(64)-\textup{SubModule}(128)$ for encoder, $\textup{SubModule}(128)\times3$ for JDSR module, and $\textup{SubModule}(128)\times2$ for TFBFI module. The decoder is composed of a $\textup{SubModule}(64)$ and a convolutional layer with the output channels of $3\cdot r^{2}$. For all layers, we use $3\times3$ kernel with a stride of 1. In addition, our method was implemented in the \textit{PyTorch} framework \cite{paszke2017pytorch} with an Adam optimizer \cite{kingma2014adam}, and an Intel Xeon E5 CPU and an NVIDIA Titan Xp GPU were utilized to train the network.

\section{Experimental Results}
\label{sec: Experimental Results}

To evaluate the performance of the proposed method on the joint video enhancement task, we compared the proposed method with the state-of-the-art (SOTA) video enhancement techniques. Because there were few methods that jointly performed three video enhancement tasks concurrently, we constructed several sequential approaches that consisted of multiple video enhancement methods and improved the quality of video sequentially. To reduce motion blurs on the input frames, SOTA VD algorithms, such as CDVD \cite{42pan2020cascaded} and STFAN \cite{40zhou2019spatio}, were used. To improve the resolution of the video, we employed SOTA video enhancement methods: Super-SloMo \cite{17jiang2018super} and AdaCoF \cite{19lee2020adacof} for VFI; and RBPN \cite{6haris2019recurrent} for VSR. In addition, recent joint video enhancement methods were also implemented, which were STAR \cite{44haris2020starnet} and Zooming-SloMo \cite{45xiang2020zooming} for VSR and VFI; and BIN \cite{43shen2020blurry} for VD and VFI. We re-trained CDVD \cite{42pan2020cascaded} and STFAN \cite{40zhou2019spatio} with our Adobe240 \cite{36su2017deep} training dataset for fair comparisons. In our experiments, we evaluated the performance of the proposed method and other sequential approaches on two test datasets: Adobe240 \cite{36su2017deep} and GOPRO \cite{GOPRO_Nah_2017_CVPR}.

\subsection{Quantitative Evaluations}

The blurry frames in the test datasets were reduced to $1/4$ size by bicubic filtering and fed into the proposed method and the sequential approaches. Then, they produced the deblurred and up-scaled frames corresponding to the keyframes and the intermediate frames. To quantitatively measure the performance of the approaches, we evaluated the peak signal-to-noise ratio (PSNR) and structural similarity index (SSIM) between the output frame and the ground-truth frame. The average PSNR and SSIM values of the Adobe240 \cite{36su2017deep} test set are presented in Table~\ref{table:1quanAdobe}. 

\begin{table*}[t]
\caption{Quantitative comparison with other sequential approaches in terms of PSNR and SSIM on Adobe240 \cite{36su2017deep} dataset.}
\begin{center}
\begin{tabular}{lcccccccc}
\hlineB{2.0}
\multirow{2}{*}{Method} & \multicolumn{2}{c}{\begin{tabular}[c]{@{}c@{}}Deblurred\\ (input scale)\end{tabular}} & \multicolumn{2}{c}{\begin{tabular}[c]{@{}c@{}}Deblurred \\ and up-scaled\end{tabular}} & \multicolumn{2}{c}{\begin{tabular}[c]{@{}c@{}}Deblurred, \\ up-scaled, and \\ interpolated\end{tabular}} & \multicolumn{2}{c}{\begin{tabular}[c]{@{}c@{}}Enhanced video\\ $(4\times)$\end{tabular}} \\ \cline{2-9} 
                        & PSNR                                       & SSIM                                     & PSNR                                       & SSIM                                      & PSNR                                                & SSIM                                               & PSNR                                        & SSIM                                       \\ \hlineB{2.0}

CDVD+STAR               & 32.717                                     & 0.955                                    & 26.737                                     & 0.893                                     & 26.162                                              & 0.885                                              & 26.450                                      & 0.889                                      \\
CDVD+Zooming-SloMo      & 32.717                                     & 0.955                                    & 26.901                                     & 0.896                                     & 26.507                                              & 0.890                                              & 26.704                                      & 0.893                                      \\
Zooming-SloMo+STFAN     & -                                          & -                                        & 26.997                                     & 0.891                                     & 26.730                                              & 0.888                                              & 26.863                                      & 0.890                                      \\
STAR+STFAN              & -                                          & -                                        & 27.964                                     & 0.908                                     & 27.058                                              & 0.895                                              & 27.511                                      & 0.902                                      \\
RBPN+STFAN+AdaCoF       & -                                          & -                                        & 28.325                                     & 0.914                                     & 27.047                                              & 0.900                                              & 27.686                                      & 0.907                                      \\
RBPN+STFAN+Super-SloMo  & -                                          & -                                        & 28.325                                     & 0.914                                     & 27.060                                              & 0.900                                              & 27.692                                      & 0.907                                      \\
BIN+RBPN                & 34.849                                     & 0.970                                    & 27.750                                     & 0.900                                     & 27.838                                              & 0.901                                              & 27.794                                      & 0.901                                      \\
STFAN+STAR              & 35.040                                     & \textbf{0.971}                           & 28.025                                     & 0.906                                     & 26.996                                              & 0.895                                              & 27.510                                      & 0.901                                      \\
STFAN+Zooming-SloMo     & 35.040                                     & \textbf{0.971}                           & 28.214                                     & 0.909                                     & 27.464                                              & 0.902                                              & 27.839                                      & 0.906                                      \\
Proposed (single)       & 34.633                                     & 0.963                                    & 28.575                                     & 0.919                                     & 28.473                                              & 0.920                                              & 28.524                                      & 0.920                                      \\
Proposed                & \textbf{35.864}                            & 0.968                                    & \textbf{29.492}                            & \textbf{0.932}                            & \textbf{29.361}                                     & \textbf{0.932}                                     & \textbf{29.427}                             & \textbf{0.932}                             \\ \hlineB{2.0}
\end{tabular}
\end{center}
\label{table:1quanAdobe}
\end{table*}

The performance of the entire enhanced video is denoted in the `Enhanced video (4$\times$)' column in Table~\ref{table:1quanAdobe}. As shown in this column, our approach achieved the best performance on the joint video enhancement task. The average PSNR and SSIM values of our approach were much higher than those of the other sequential approaches. These results demonstrate that the sequential approaches are sub-optimal and have several limitations. Because the proposed method jointly treats three video enhancement tasks with the integrated enhancement model, it outperforms the other methods.

For all sequential approaches except BIN+RBPN, the metric values (i.e. PSNR and SSIM) of the intermediate frames are much lower than those of the deblurred and up-scaled frames. It means that other degradation factor, i.e. motion blur in this case, disturbs the frame interpolation. Even though SOTA video deblurring methods can reduce motion blurs in the sequential approaches, the remained motion blur may degrade the performance of the frame interpolation. BIN+RBPN and the proposed method could produce the deblurred and up-scaled frame and the intermediate frame with similar performance because they handled both video deblurring and frame interpolation jointly.

Our DSFN directly produces the enhanced frames in the $4\times$ up-scaled size. The deblurred frames in the scale of the inputs (i.e. $1/4$ scale of the original frames) can be simply obtained by down-sampling the outputs with bicubic filtering. Then we could measure the performance of the deblurred outputs in the scale of the inputs. In accordance with the `Deblurred (input scale)' column in Table~\ref{table:1quanAdobe}, the proposed method achieved the best performance on the deblurred outputs in the input scale in terms of PSNR. Meanwhile, STFAN recorded a similar performance to the proposed method in terms of SSIM. However, the performances of the final enhanced outputs of the sequential methods (STFAN+STAR and STFAN+Zooming-SloMo) are much lower than those of our approach. This is because the proposed method could obtain more optimal results by resolving multiple degradation factors simultaneously. The superior performance of the proposed method on the joint video enhancement task is also demonstrated in GOPRO dataset \cite{GOPRO_Nah_2017_CVPR}, which is shown in Table~\ref{table:2quanGOPRO}.

\begin{table}[t]
\caption{Quantitative comparison with other sequential approaches in terms of PSNR and SSIM on GOPRO \cite{GOPRO_Nah_2017_CVPR} dataset.}
\vspace{-0.3cm}
\begin{center}
\begin{tabular}{lcccc}
\hlineB{2.0}
\multirow{2}{*}{Method} & \multicolumn{2}{c}{\begin{tabular}[c]{@{}c@{}}Deblurred \\ and up-scaled\end{tabular}} & \multicolumn{2}{c}{\begin{tabular}[c]{@{}c@{}}Deblurred, \\ up-scaled, and \\ interpolated\end{tabular}} \\ \cline{2-5} 
                        & PSNR                                       & SSIM                                      & PSNR                                                & SSIM                                               \\ \hlineB{2.0}
CDVD+STAR               & 26.409                                     & 0.878                                     & 25.401                                              & 0.855                                              \\
CDVD+Zooming-SloMo      & 26.483                                     & 0.880                                     & 25.808                                              & 0.865                                              \\
STFAN+STAR              & 27.273                                     & 0.886                                     & 26.017                                              & 0.864                                              \\
STFAN+Zooming-SloMo     & 27.352                                     & 0.888                                     & 26.472                                              & 0.873                                              \\
BIN+RBPN                & 27.527                                     & 0.888                                     & 27.596                                              & 0.890                                              \\
Proposed (single)       & 27.292                                     & 0.890                                     & 27.362                                              & 0.896                                              \\
Proposed                & \textbf{28.259}                            & \textbf{0.909}                            & \textbf{28.254}                                     & \textbf{0.911}                                     \\ \hlineB{2.0}
\end{tabular}
\end{center}
\label{table:2quanGOPRO}
\end{table}


\subsubsection{Effectiveness of the JDSR Module}

The performance of the deblurred and up-scaled frames is denoted in the `Deblurred and up-scaled' column in Table~\ref{table:1quanAdobe}. According to this column, the proposed method also outperforms the other sequential approaches on the deblurring and super-resolution task. In the proposed method, these deblurred and up-scaled frames can be obtained by the JDSR module. Therefore, we could confirm that the proposed JDSR performs well both deblurring and super-resolution in the DSFN. 

\subsubsection{Effectiveness of the TFBFI Module}

The `Deblurred, up-scaled, and interpolated' column in Table~\ref{table:1quanAdobe} represents the performance of the intermediate frames between the input frames. This column shows that the proposed method generates the intermediate frame most similar to the original frame. Because the proposed TFBFI module appropriately warped keyframes to the intermediate frames with abundant information from three successive frames, the DSFN was able to obtain better results than the other VFI methods.

\subsection{Qualitative Evaluations}

\begin{figure*}
\begin{center}
\includegraphics[width=\linewidth]{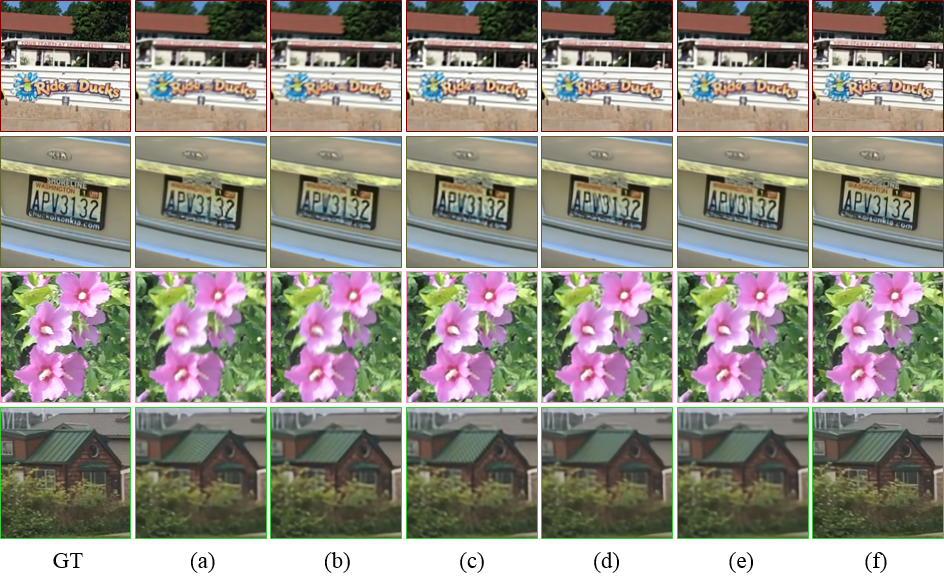}
\end{center}
\vspace{-0.5cm}
   \caption{Deblurred and up-scaled intermediate frames from various methods on Adobe240 \cite{36su2017deep} and GOPRO \cite{GOPRO_Nah_2017_CVPR} dataset. The various methods are (a) CDVD+STAR, (b) CDVD+Zooming-SloMo, (c) BIN+RBPN, (d) STFAN+STAR, (e) STFAN+Zooming-SloMo, and (f) the proposed method.}
\label{fig:qual}
\end{figure*}

To verify the superiority of the proposed method on the joint video enhancement task, examples of the output frames from the proposed method and other sequential methods are shown in Figure~\ref{fig:qual}. As shown in Figure~\ref{fig:qual}, the proposed method produced most similar frames to the ground-truth in comparison to the other sequential approaches. Especially, the sequential methods generated the blurry output frames while the proposed method effectively removed motion blurs on the inputs. These results show that the proposed DSFN could solve the joint video enhancement problem elaborately. 

\begin{figure*}
\begin{center}
\includegraphics[width=\linewidth]{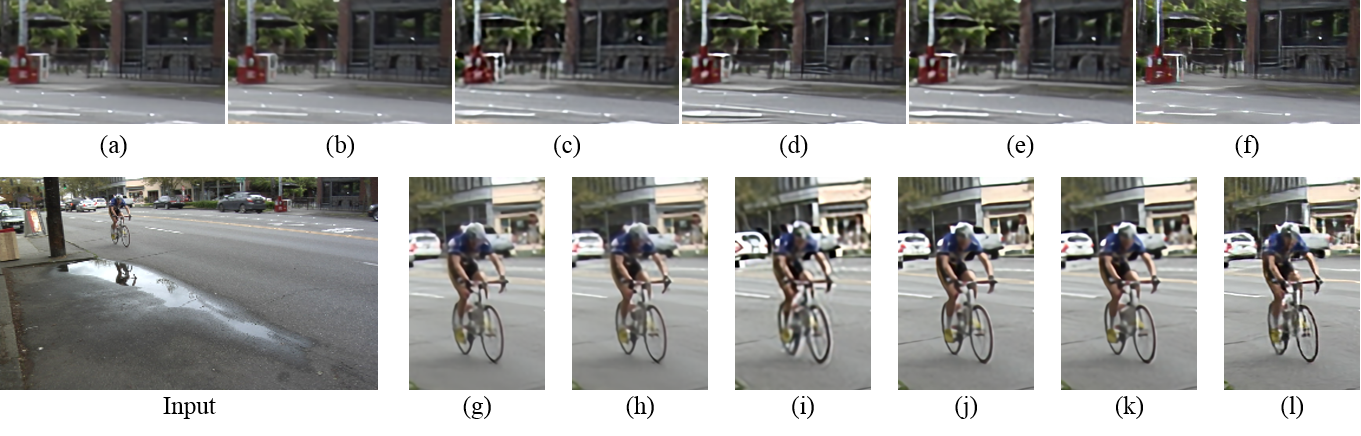}
\end{center}
\vspace{-0.5cm}
   \caption{Joint video enhanced frames from various methods on a real blurry dataset \cite{26cho2012video}. The various methods are (a), (g) CDVD+STAR, (b), (h) CDVD+Zooming-SloMo, (c), (i) BIN+RBPN, (d), (j) STFAN+STAR, (e), (k) STFAN+Zooming-SloMo, and (f), (l) the proposed method.}
\label{fig: sup1}
\end{figure*}

\begin{figure*}
\begin{center}
\includegraphics[width=\linewidth]{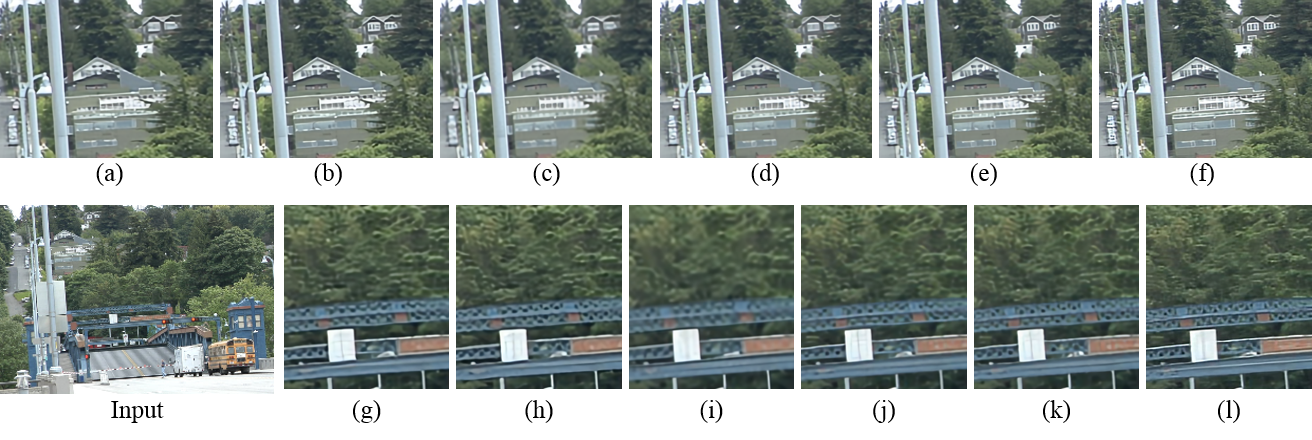}
\end{center}
\vspace{-0.5cm}
   \caption{Joint video enhanced frames from various methods on a real blurry dataset \cite{26cho2012video}. The various methods are (a), (g) CDVD+STAR, (b), (h) CDVD+Zooming-SloMo, (c), (i) BIN+RBPN, (d), (j) STFAN+STAR, (e), (k) STFAN+Zooming-SloMo, and (f), (l) the proposed method.}
\label{fig: sup2}
\end{figure*}

In addition, we applied the proposed method and other existing methods to a real blurry dataset from \cite{26cho2012video}. In this experiment, the proposed method can also handle real-world blurry data properly even though the proposed network, DSFN, was trained by the synthetic dataset \cite{36su2017deep}. As illustrated in Figures~\ref{fig: sup1}~and~\ref{fig: sup2}, we can confirm that the proposed method generates the enhanced frames well compared to the other sequential methods.

\subsection{Network Size and Processing Time}

We measured the network size and the processing time of each approach on the joint video enhancement task. To accurately measure the processing time, the inference time of processing 30 frames was measured and averaged in our environment. We utilized \textit{torch.cuda.synchronize()} function to wait for all kernels in all streams on a CUDA device to complete. As shown in Table~\ref{table:3size}, the proposed method effectively performed the joint enhancement task, whereas the network size is small and the processing time is fast. This is because input frames are jointly deblurred and super-resolved by JDSR module while other approaches perform deblurring and super-resolution operations sequentially. In addition, because the appropriate feature maps are extracted in advance from the encoder and they are utilized in both JDSR and TFBFI modules, we can prevent redundant layers in DSFN. The proposed method can efficiently produce the enhanced outputs since all tasks are conducted through the feature maps in the input scale.

\begin{table}[t]
\caption{Network size and processing time of each approach.}
\vspace{-0.3cm}
\begin{center}
\begin{tabular}{lcc}
\hlineB{1.5}
\multirow{2}{*}{Method} & Parameters    & Average runtime \\
                        & (million)     & (s/frame)       \\ \hlineB{1.5}
CDVD+STAR               & 16.2+11.2     & 0.81            \\
CDVD+Zoom               & 16.2+11.1     & 0.74            \\
RBPN+STFAN+AdaCoF       & 12.7+5.4+21.8 & 0.73            \\
RBPN+STFAN+SloMo        & 12.7+5.4+19.8 & 6.12            \\
STFAN+STAR              & 5.4+11.2      & 0.36            \\
STFAN+Zoom              & 5.4+11.1      & 0.29            \\
BIN+RBPN                & 4.7+12.7      & 0.89            \\
Proposed                & \textbf{10.9} & \textbf{0.02}   \\ \hlineB{1.5}
\end{tabular}
\end{center}
\label{table:3size}
\end{table}

\subsection{Visualization of the Offset Fields}

\begin{figure*}
\begin{center}
    \includegraphics[width=\linewidth]{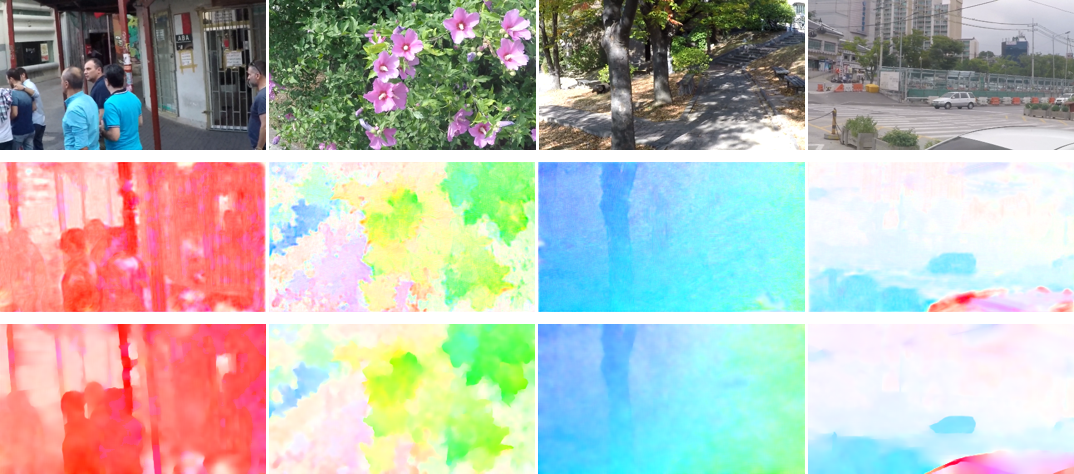}
\end{center}
\vspace{-0.3cm}
   \caption{Input frames (the first row), their corresponding visualized offset fields (the second row), and their corresponding visualized optical flows from \cite{hui2018liteflownet} (the third row).}
\label{fig: visualization}
\end{figure*}

In the proposed TFBFI module, the offset fields which infer object motions between the input adjacent frames are estimated as in Eq.~\ref{two_off}. To verify how well the estimated offset fields represent object motions, we visualized them as shown in Figure~\ref{fig: visualization}. Because we utilized deformable convolution with $3\times3$ kernel, there are nine offset values for each location. We averaged these offset values and visualized them as if they were optical flows. These results were then compared with optical flow maps which were estimated by \cite{hui2018liteflownet}. According to Figure~\ref{fig: visualization}, the visualized offset fields are similar to the estimated flow maps from \cite{hui2018liteflownet}. This means that the estimated offset field appropriately represents the object motions between the input adjacent frames.

\section{Conclusion}
\label{sec: Conclusion}

In this paper, we proposed the novel joint video enhancement method that performed video deblurring, super-resolution, and frame interpolation, simultaneously. To jointly handle multiple video enhancement tasks, we newly designed DSFN that produced the deblurred and up-scaled frames for input frames and the intermediate frames at the same time. The enhanced outputs could be effectively obtained through our DSFN because it resolved the integrated enhancement problem with the proper combination of the proposed JDSR and TFBFI modules. In our experiments, the proposed method outperformed the other sequential approaches that consisted of various SOTA video enhancement techniques for the joint video enhancement task. While achieving superior performance, our approach allowed to perform joint enhancement more efficiently with smaller network size and faster processing time.

\bibliographystyle{unsrt}  
\bibliography{references}

\end{document}